\documentclass{article}

\usepackage{neurips_2022}

\usepackage[utf8]{inputenc} 
\usepackage[T1]{fontenc}    
\usepackage{hyperref}       
\usepackage{url}            
\usepackage{booktabs}       
\usepackage{amsfonts}       
\usepackage{nicefrac}       
\usepackage{microtype}      
\usepackage{xcolor}         

\title{Finetuning for Sarcasm Detection with a Pruned Dataset}

\author{%
  Ishita Goyal \thanks{Equal contribution}\\
  School of Computer Science\\
  Carnegie Mellon University\\
  Pittsburgh, PA 15213 \\
  \texttt{igoyal@andrew.cmu.edu} \\
  \And
  Priyank Bhandia \footnotemark[1]\\
  School of Computer Science\\
  Carnegie Mellon University \\
  Pittsburgh, PA 15213 \\
  \texttt{pbhandia@andrew.cmu.edu} \\
  \AND
  Sanjana Dulam \footnotemark[1]\\
  School of Computer Science\\
  Carnegie Mellon University \\
  Pittsburgh, PA 15213 \\
  \texttt{sdulam@andrew.cmu.edu} \\
}

\begin{document}

\maketitle

\begin{abstract}
  Sarcasm is a form of irony that involves saying or writing something that is opposite or opposite to what one really means, often in a humorous or mocking way. It is often used to mock or mock someone or something, or to be humorous or amusing. Sarcasm is usually conveyed through tone of voice, facial expressions, or other forms of nonverbal communication, but it can also be indicated by the use of certain words or phrases that are typically associated with irony or humor. Sarcasm detection is difficult because it relies on context and non-verbal cues. It can also be culturally specific, subjective and ambiguous. In this work, we fine-tune the RoBERTa based sarcasm detection model presented in \citet{abaskohi2022utnlp} to get to within 0.02 F1 of the state-of-the-art (\citet{hercog2022sarcastic}) on the iSarcasm dataset (\citet{oprea2019isarcasm}). This performance is achieved by augmenting iSarcasm with a pruned version of the Self Annotated Reddit Corpus (SARC) (\cite{khodak2017large}). Our pruned version is 100 times smaller than the subset of SARC used to train the state-of-the-art model. Our code is available at: \url{https://github.com/priyank96/dataset-pruning-sarcasm-detection}
\end{abstract}

\section{Introduction}
In this work we try to reduce the amount of data needed to achieve near state-of-the-art performance on the sarcasm detection task. Our chosen datasets consist of short texts (iSarcasm - tweets and SARC - comments under reddit posts).
We apply a data pruning technique to reduce the train dataset size. We also experiment with different loss functions. We are motivated to apply dataset pruning to reduce training time, which has the benefits of reducing the time needed to perform hyperparameter search, consuming less energy, and so on. We are motivated to experiment with different loss functions from the observation that the SARC dataset is noisier than the iSarcasm dataset and so more robust loss functions could yield better performance (i.e, higher F1 scores).  
\section{Related Work}
We choose to fine tune the RoBERTa based sarcasm detection model proposed in \citet{abaskohi2022utnlp}. The authors experiment with a range of models (SVM-based, BERT-based, Attention-based and LSTM-based), as well as different data augmentation strategies (generative-based and mutation-based). In an effort to increase the size of the train set, they also try to augment the iSarcasm dataset with the Sentiment140 dataset (\citet{go2009twitter}) and the Sarcasm Headlines Dataset (\citet{misra2019sarcasm}). Crucially, they do not augment with SARC, our chosen dataset. The authors find that RoBERTa, fine tuned for sentiment classification, without any data augmentation and further fine tuned only on the iSarcasm dataset gives them the best performance, a F1 score of 0.414.
This model falls short of the same RoBERTa model finetuned for sarcasm detection by \citet{hercog2022sarcastic}, using a training set comprising iSarcasm and a subset of SARC to obtain a F1 score of 0.526 on iSarcasm. 
\\
The iSarcasm dataset (\citet{oprea2019isarcasm}) is a dataset of intended sarcasm, in the sense that each of the tweets in this dataset is labeled as sarcastic and non-sarcastic by the author of the tweet themselves. The labeling of intended sarcasm makes iSarcasm unique, and the authors show that previously state-of-the-art models trained on datasets that label perceived sarcasm perform poorly on iSarcasm. iSarcasm consists of 777 sarcastic and 3,707 non-sarcastic tweets. \\
The Self Annotated Reddit Corpus (\cite{khodak2017large}) is a dataset of Reddit comments labeled for perceived sarcasm. They label comments using distant supervision - comments containing the "/s" token are considered as sarcastic. This comes from the convention among Reddit users to signal sarcasm with "/s" at the end of a comment. Still, SARC is noisy as this is a convention not a law. The authors claim that 2\% of the samples in SARC are false negatives (sarcastic, but do not have “/s”) and 1\% of the samples are false positives (not sarcastic, but contain “/s”). To put this in context, the full dataset has only 1\% sarcastic comments. We use the balanced subset of SARC used by \cite{hercog2022sarcastic} as our starting point and attempt to prune it to make it smaller while maintaining similar performance. To do the dataset pruning, we turn to the student-teacher setting for perceptron learning described by \citet{sorscher2022beyond}.\\
In \citet{sorscher2022beyond}, the authors look for ways to select training samples such that the reduction in test error does better than the scaling laws, described in \cite{kaplan2020scaling} and \cite{hoffmann2022training}, which tell us to expect power law scaling - a drop in error from 3\% to 2\% may need an order of magnitude more data. To escape this, careful selection of training data is prescribed. The authors show both theoretically and experimentally, that for their chosen ImageNet dataset: 
“(a) The optimal pruning strategy changes depending on the amount of initial data; with
abundant (scarce) initial data, one should retain only hard (easy) examples.
(b) Exponential scaling is possible with respect to pruned dataset size provided one chooses
an increasing Pareto optimal pruning fraction as a function of initial dataset size.” They show that these results hold not only for training from scratch but also for finetuning (our setting).\\
We also experiment with different loss functions and study their effect on task performance. \citet{abaskohi2022utnlp} use the Cross Entropy Loss. To address the class imbalance in the iSarcasm dataset, a natural thought would be to use Weighted Cross Entropy Loss. One of the drawbacks of this approach is that by up-weighting the minority class, there is a risk that the number of false positives may increase. 
Squared Hinge Loss has been shown to perform well in classification tasks (\citet{lee2015deeply}). In \citet{tang2013deep}, the authors show that a well fitted Hinge Loss can outperform log loss based networks. 
Finally, in \citet{janocha2017loss}, the authors attempt to provide a comparison of various losses under various criteria and advise on when (and why) to use them. From one of their experiments (measuring speed of learning expressed as expected train/test accuracy for different dataset sizes), they note that squared hinge loss converges faster than log losses - a property that is potentially useful to us because we may be operating in a data scarce regime (iSarcasm is a small dataset!). 
The experimental results in \citet{janocha2017loss} also show evidence that squared hinge loss is more robust to noise in the training set labeling space as well as slightly more robust to noise in the input space. Again, this is a property that can be helpful to us when we augment with SARC, which is noisier than the iSarcasm dataset by fact of being labeled using distance supervision. 
\section{Experiments}
For our experiments, we use the twitter-RoBERTa-sentiment-base model available at \url{https://huggingface.co/cardiffnlp/twitter-roberta-base-sentiment}, with the following hyperparameters:\\
Batch Size: 32\\
Weight Decay: 0.01\\
Warmup Steps: 500\\
Epochs: 5
\subsection{Data Augmentation and Data Pruning}
We use the student-teacher setting for perceptron learning technique described in \citet{sorscher2022beyond}. This approach involves using a teacher model to label points from the dataset, and then keeping only the points which were misclassified with high (low) confidence, if your teacher model was trained on abundant (scarce) data. 
For our setting, we take the initial baseline model trained on the iSarcasm dataset as the teacher model to classify training samples from SARC. We then order the teacher model’s mis-classifications by confidence (logit score). We say that an example mis-classified with low confidence (logit score) is an easy example, and an example mis-sclassified with high confidence (logit score) is a hard exmaple.\\  
From these ordered data points, we form three new datasets: iSarcasm + 10,000 SARC train examples mis-classified with lowest confidence (Easy SARC Examples), iSarcasm + 10,000 SARC train examples mis-classified with highest confidence (Hard SARC Examples) and iSarcasm + 10,000 random SARC examples. The model is then fine tuned on these new datasets and results are reported in \autoref{tab:table1}.\\ We note that by taking the dataset with the easy SARC examples, we achieve results better than the baseline. This aligns with the findings put forward in \citet{sorscher2022beyond}, “The optimal pruning strategy changes depending on the amount of initial data; with
abundant (scarce) initial data, one should retain only hard (easy) examples.” In our case, the iSarcasm dataset is small so we have scarce initial data, and we see that the model performance improves by adding easy examples to the dataset. 
\begin{table}[h]
\centering
\begin{tabular}{@{}cc@{}}
\toprule
Dataset                         & F1 Score        \\ \midrule
iSarcasm                        & 0.4031          \\ \midrule
iSarcasm + Hard SARC Examples   & 0.2663          \\ \midrule
iSarcasm + Easy SARC Examples   & \textbf{0.4507} \\ \midrule
iSarcasm + Random SARC Examples & 0.3150          \\ \bottomrule
\end{tabular}
\caption{F1 Score for Different Datasets}
\label{tab:table1}
\end{table}

\subsection{Loss Functions}
The model in \citet{abaskohi2022utnlp} was finetuned using Cross Entropy Loss. As our chosen dataset is imbalanced (the count of sarcastic tweets is roughly 1/5th of the count of non sarcastic tweets), first we use Weighted Cross Entropy Loss. We assign a weight of 0.75 to the sarcastic class and a weight of 0.25 to the non sarcastic class. Next, we experiment with Hinge loss and Squared Hinge Loss. F1 Scores on the test set and training losses are reported in \autoref{tab:table2}. The training was done on the un-augmented iSarcasm dataset.\\
From this experiment we observed that all the alternative loss functions achieved a greater F1 score than the Cross Entropy. Weighted Cross Entropy out performs both the hinge losses.
\begin{table}[h]
\centering
\begin{tabular}{@{}cc@{}}
\toprule
Loss Function          & F1 Score \\ \midrule
Cross Entropy          & 0.4031   \\ \midrule
Weighted Cross Entropy & \textbf{0.4410}   \\ \midrule
Hinge Loss             & 0.4341   \\ \midrule
Squared Hinge Loss     & 0.4114   \\ \bottomrule
\end{tabular}
\caption{F1 Score for Different Loss Functions}
\label{tab:table2}
\end{table}
\subsection{Loss Function and Data Pruning}
For the final experiment, we combine the previous two experiments to arrive at the final model for sarcasm detection that our work recommends. Results are reported in \autoref{tab:table3}.\\
The combination of Squared Hinge Loss with the iSarcasm + Easy SARC Examples dataset achieves a F1 score of 0.5067, which is only 0.02 away from the state-of-the-art F1 score of 0.526, achieved in \citet{hercog2022sarcastic} by augmenting iSarcasm with the full SARC dataset. So, we achieve our F1 score by augmenting with 100 times less data (ten thousand rows from SARC instead of 1 million).
\begin{table}
\begin{tabular}{@{}cccc@{}}
\toprule
 & \begin{tabular}[c]{@{}c@{}}iSarcasm +\\  Hard SARC Examples\end{tabular} & \begin{tabular}[c]{@{}c@{}}iSarcasm + \\ Easy SARC Examples\end{tabular} & \begin{tabular}[c]{@{}c@{}}iSarcasm + \\ Random SARC Examples\end{tabular} \\ \midrule
Cross Entropy & 0.2663 & 0.4507 & 0.3150 \\ \midrule
Weighted Cross Entropy & 0.1638 & 0.4750 & 0.2377 \\ \midrule
Hinge Loss & 0.2933 & 0.4718 & 0.2488 \\ \midrule
Squared Hinge Loss & 0.2956 & \textbf{0.5067} & 0.3154 \\ \bottomrule
\end{tabular}
\caption{ F1 Scores of Different Loss Functions over Different Datasets}
\label{tab:table3}
\end{table}
\section{Conclusion}
In this work, we saw that by applying dataset pruning it is possible to achieve performance comparable to models trained on much larger training data. By ordering misclassified samples by classification confidence (logit scores), we arrive at a difficulty ordering of the data, which makes it possible to apply curriculum learning (\citet{bengio2009curriculum}) to our task, which could give further performance improvements and is left as future work.
\bibliographystyle{unsrtnat}
\bibliography{bibliography}
\end{document}